\documentclass{article}
\usepackage{lipsum}

\usepackage{spconf,amsmath,graphicx}
\usepackage{microtype}
\usepackage{array}
\newcolumntype{P}[1]{>{\centering\arraybackslash}p{#1}}
\newcolumntype{M}[1]{>{\centering\arraybackslash}m{#1}}

\title{HOLISTIC FEATURES FOR REAL-TIME CROWD BEHAVIOUR ANOMALY DETECTION}
\name{ Mark Marsden   \: \:      Kevin McGuinness \: \: Suzanne Little \: \:  Noel E. O'Connor}
\address{Insight Centre for Data Analytics \\  Dublin City University \\ Dublin, Ireland}

\begin{document}
%
\maketitle
\begin{abstract}

This paper presents a new approach to crowd behaviour anomaly detection that uses a set of efficiently computed, easily interpretable, scene-level  holistic features. This low-dimensional descriptor combines two features from the literature:  crowd collectiveness \cite{zhou2014measuring} and crowd conflict \cite{Kang2015}, with two newly developed crowd features: mean motion speed and a new formulation of crowd density. Two different anomaly detection approaches are investigated using these features. When only normal training data is available we use a Gaussian Mixture Model (GMM) for outlier detection. When both normal and abnormal training data is available we use a Support Vector Machine (SVM) for binary classification. We evaluate on two crowd behaviour anomaly detection datasets,  achieving both state-of-the-art classification performance on the violent-flows dataset \cite{Hassneretal:SISM12} as well as better than real-time processing performance (40 frames per second).

\end{abstract}
\section{Introduction}
\label{sec:intro}

The unprecedented rise in CCTV surveillance in the last decade has led to more video data being produced than can be analysed by a human observer. Therefore, automated, real-time analysis of these ever growing archives has become a key challenge for the computer vision community. Real-time computational performance is vitally important in large-scale surveillance scenarios where scalability and a rapid response time are needed.

Crowd behaviour anomaly detection has been the focus of much research in recent years.  Anomalous crowd behaviour is that which strays significantly from an established norm, which is typically learned from ``normal'' training examples.  Abnormal crowd behaviour can correspond to (but is not limited to) panic events as well as violent and antisocial behaviour. An outlier detection strategy is usually taken when there is a  limited amount of ``abnormal'' training data available. Crowd \textit{behaviour} anomaly detection focuses on the motion characteristics, interactions, and spatial distribution within a crowd system and should not be confused with  \textit{visual} anomaly detection (e.g. an unexpected vehicle entering a pedestrian area). Most of the work to date in this area can be divided into two categories, depending on the low-level features analysed: \textit{i)} spatio-temporal volumes (STVs) and \textit{ii)} optical flow/tracklets.

Approaches which analyse spatio-temporal volumes include Mixture of Dynamic Textures \cite{Li2014a}  and Spatio-Temporal Compositions \cite{JavanRoshtkhari2013}. These techniques divide a video sequence into a set of spatio-temporal cuboids of pixels before classifying each as abnormal or not by comparing to a trained model.  This type of approach allows for both abnormal behaviour and visual anomalies to be detected but tends to operate far below real-time performance ($< 5$ frames per second) \cite{JavanRoshtkhari2013}.
 
Techniques based on optical flow/tracklets include the Social Force Model~\cite{mehran2009abnormal}, HOT (Histogram of Oriented Tracklets)~\cite{mousavi2015analyzing}, Substantial Derivative~\cite{Mohammadi2015}, Commotion Measure~\cite{mousavi2015crowd} and ViF (violent-flows descriptor)~\cite{Hassneretal:SISM12}. These approaches collate local motion vectors before applying a physics inspired model to  generate a representation of crowd behaviour which can be used to classify crowd behaviour. These approaches tend to focus solely on the detection of anomalous behaviour and are generally more computationally efficient.
 
Our approach falls into the latter category, analysing tracklet information to produce a low-dimensional descriptor of scene-level, easily interpretable, holistic crowd features.  Two holistic crowd properties from the literature, crowd conflict \cite{Kang2015} and crowd collectiveness \cite{zhou2014measuring}, are combined with two newly developed features, mean motion speed and a new formulation of crowd density, to produce a 4 dimensional descriptor of crowd behaviour that can be extracted in real-time. 

Training data for abnormal crowd events is scarce due to their infrequent occurrence in the real world. This results in crowd behaviour models often being trained using purely normal behaviour training data. When only normal training data is available we use a Gaussian Mixture Model (GMM) for outlier detection. When both normal and abnormal training data is available we use a Support Vector Machine (SVM) for binary classification.

The paper is structured as follows. Section~\ref{sec:holistic} presents a technical explanation of the holistic features used. Section~\ref{sec:anomaly} discusses the two detection approaches employed. Section~\ref{sec:experiments} describes the experimental results obtained on two commonly used crowd behaviour anomaly datasets.

\section{Holistic Features}
\label{sec:holistic}

This section provides a technical description of the tracklet and feature extraction stages. The objective is to design a low dimensional set of features that are quick to compute and capture sufficient holistic information about the objects moving in a scene to allow straightforward discrimination between normal and abnormal events. With this in mind, we propose four such features that we later show can be used effectively to detect abnormal events. Two of these features, crowd collectiveness and crowd conflict, attempt to model the interaction of objects in crowded scenes and are based on previous work in crowd analysis \cite{zhou2014measuring,Kang2015}. The second two features are intended to model crowd density and motion, which we found significantly improve classification performance without requiring significant additional computation.

\subsection{Tracket Extraction} Tracklet extraction is performed by first segmenting the image foreground using the Gaussian-mixture based method of KaewTraKulPong and Bowden~\cite{kaewtrakulpong2002improved}, after which a morphological opening and closing is performed to remove isolated noise. This results in only the dynamic parts of the scene being tracked. Interest points within the segmented foreground are then found using the selection method of Shi and Tomasi \cite{shi1994good} and tracked using a sliding window KLT tracker \cite{tomasi1991detection}, producing a set of tracklets for a given video sequence. A new set of interest points is found every 30 frames (KLT tracked points are often lost by this stage), resulting in a maximum tracklet length of 30 frames.  Any tracklet shorter than 5 frames or found to be largely static is discarded as noise. Static tracklets are judged to be those with a low standard deviation ($< 0.1$) in their sets of of $x$ and $y$ positions. These threshold values are deemed to be reasonable cut-offs for tracklet length and activity level when trying to remove noise. Velocity vectors are calculated along each tracklet by calculating the inter-frame position differences in both the $x$ and $y$ directions.

\subsection{Feature Extraction}
Using the set of tracklets and tracklet velocities extracted from a given video sequence our 4 features can be calculated for each frame. For low-activity frames ($< 10$ currently tracked points) no features are extracted and the frame is classified as normal. 10 or more currently tracked points is seen to represent a genuine crowded scene with multiple bodies interacting. An exponential moving average filter is applied to the 1-D signal extracted for each feature, with the smoothing factor set to $0.1$. This $0.1$ smoothing factor results in significant noise removal and the core trend of each feature being extracted.

\subsubsection{Crowd Collectiveness}
Crowd collectiveness is a scene-independent holistic property of a crowd system.  It can be defined as the degree to which individuals in a scene move in unison \cite{zhou2014measuring}. Zhou et al.'s~\cite{zhou2014measuring} method for measuring this property analyses the tracklet positions and velocities found in the current frame and constructs a weighted adjacency matrix. The edge weights within each matrix column are summed and the mean is calculated. This mean value corresponds to the overall collectiveness level for the current frame. 

\subsubsection{Crowd Conflict}
Crowd conflict is another scene-independent holistic crowd property which can be defined as the level of friction/interaction between neighbouring tracked points \cite{shao2014scene}.  Shao et al.~\cite{Kang2015} efficiently calculate this property by summing the velocity correlation between each pair of neighbouring tracked points in a given frame.

\subsubsection{Crowd Density}
\begin{figure}[t]
	\centering
	\includegraphics[width=0.25\textwidth]{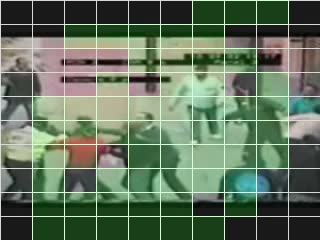}
	\caption{Crowd density calculation grid for a scene from the violent-flows dataset. Each green square corresponds to an occupied grid cell (crowd density in this frame = 57\%).}
    \label{bli}
\end{figure}

Crowd density can be defined as the level of congestion observed across a scene at a given instant. Our unique approach to calculating this features divides the scene into a fixed size grid ($10 \times 10$) and counts the number of grid cells currently occupied by one of more tracked points. Equation 1 is then used to calculate the crowd density level for the current frame. A $10 \times 10$ grid was chosen to provide sufficient granularity in the density calculation, with the aim being for each grid   cell to roughly contain one or two pedestrian in most surveillance scenarios. There are obvious limitations in terms of  scale invariance with this feature, however the main objective is not pixel perfect accuracy but to measure a useful crowd property in a  highly efficient manner.  Figure~\ref{bli} shows our density feature being calculated for a frame taken from the the violent-flows dataset.

\begin{equation}
\text{Crowd Density}= \frac{\text{Occupied Grid Cells}}{\text{Total Grid Cells}}
\end{equation}

\subsubsection{Mean Motion Speed}
The mean motion speed observed within a crowded scene provides a coarse, scene-level feature that can be extracted very efficiently. Our approach estimates this crowd property by calculating the magnitude of each tracklet velocity vector in the current frame and finding the mean. While conceptually simple our experiments in Section~\ref{sec:experiments} show that the inclusion of this feature noticeably improves the accuracy of crowd behaviour anomaly detection.

\section{Anomaly Detection Approaches}
\label{sec:anomaly}

We investigate two anomaly detection approaches. Each require the following pre-processing steps. All individual feature are firstly scaled to lie within the range $[0,1]$, with respect to the range of training data values. Normalisation is then performed by dividing by the maximum magnitude vector in the training set. The low-dimensional descriptor used results in almost negligible training and classification times for reasonably sized datasets.

\subsection{Gaussian Mixture Model}
We use a Gaussian Mixture Model to perform outlier detection when only normal behaviour training data is available. The GMM configuration (number of mixture components and type of co-variance matrix) for a given experiment is selected as the one that minimizes the Bayesian Information Criterion (BIC) value~\cite{schwarz1978estimating} on the training data. The selected model is then used to calculate the log probabilities for the full set of training frames, and the distribution of these log probability values is used to decide upon an outlier detection threshold using Otsu's method \cite{otsu1975threshold}. Test frames are then classified as abnormal or not by using the fit mixture model to calculate their log probability and applying the the adaptive threshold generated from the training data.

\begin{table}[tbp]
\centering
\begin{tabular}{|M{3cm}|M{3cm}|}
\hline
No. of Mixture Components & BIC \\ \hline
1  & -20015 \\
2  &  -21810 \\ 
\textbf{3}  & \textbf{-22047}  \\ 
4 & -21940 \\ \hline
\end{tabular}
\caption{BIC values calculated during the GMM selection stage for the UMN dataset.}
\label{GMM_UMN}
\end{table}

\begin{table}[tbp]
\centering
\begin{tabular}{|M{3cm}|M{3cm}|}
\hline
No. of Mixture Components & BIC \\ \hline
1  & -51758 \\
2  &  -223161 \\ 
3  & -274742  \\ 
\textbf{4} & \textbf{-327545} \\ \hline
\end{tabular}
\caption{BIC values calculated during the GMM selection stage for the violent-flows dataset.}
\label{GMM_VF}
\end{table}

\subsection{Support Vector Machine}
We use a discriminative model (binary classifier) for outlier detection when both normal and abnormal training data are available. Specifically, we trained a Support Vector Machine with an RBF kernel on test frames labelled as normal and abnormal. The default value of 1.0 was used for the SVM regularization parameter $C$.

\section{Experiments}
\label{sec:experiments}

In this section the proposed method is evaluated on two distinct crowd behaviour anomaly datasets: \textit{i)} the UMN dataset\footnote{http://mha.cs.umn.edu/Movies/Crowd-Activity-All.avi}  and \textit{ii)} the violent-flows dataset~\cite{Hassneretal:SISM12}. These benchmarks assess the ability of a given approach to detect unusual crowd behaviour at the frame-level and the video-level respectively. All experiments were carried out using MATLAB 2014a and Python 2.7 on a 2.8 GHz Intel Core i5 processor with 8GB of RAM.

\subsection{UMN Dataset}
\begin{figure}[t]
	\centering
	\includegraphics[width=0.35\textwidth]{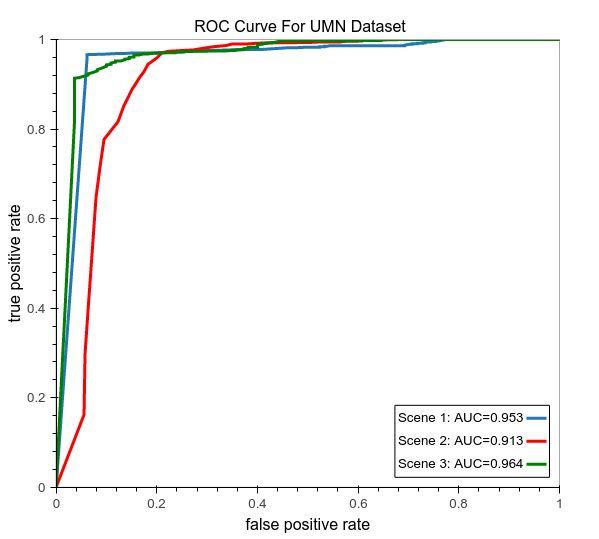}
	\caption{ROC curves for 3 UMN scenes}
    \label{3scenes}
\end{figure}

The UMN dataset contains 11 sequences filmed in 3 different locations. Each sequence begins with a period of normal passive behaviour before a panic event/anomaly occurs towards the end.  The objective here is to train a classifier using frames from the the initial normal period and evaluate it's detection performance on the subsequent test frames. Classification is performed at the frame level and results are compared in terms of ROC curve AUC (Area under the curve). For each of the three scenes the initial 200 frames of each clip are combined to form a training set, with the remaining frames used as a test set for that scene. This results in a roughly $1:2$ split between training and test frames for each camera location and will be referred to as our single scene experiment. While this dataset is quite limited in terms of size and variation it does provide a good means of performance evaluation during the development of a crowd anomaly detection algorithm.  Since no abnormal frames are made available for training in this experiment our GMM-based detection approach is used. Table \ref{GMM_UMN} presents the BIC values calculated during the GMM selection stage, with a 3 component model ultimately used. A full co-variance matrix GMM resulted in a lower BIC value in all cases and was therefore used. Figure~\ref{3scenes} presents the ROC curves for all three UMN scenes individually. A cross-scene anomaly detection approach is also taken where for a given UMN scene the training frames from the two other scenes are used to generate the GMM.

\begin{table}[tbp]
\centering
\begin{tabular}{|M{4cm}|M{1cm}|M{1.2cm}|}
\hline
Method & AUC & Speed (FPS) \\ \hline
MDT \cite{Li2014a} & 0.995 & 0.9 \\
CM \cite{mousavi2015crowd} & 0.98 & 5 \\ 
SFM \cite{mehran2009abnormal} & 0.97 & 3 \\ \hline
Our Method (Single scene) & 0.929 & \textbf{40} \\ 
Our Method (Cross-scene) & 0.869 & \textbf{40} \\ \hline
\end{tabular}
\caption{ROC curve AUC performance and processing speed on the UMN dataset.}
\label{comparison}
\end{table}

Table~\ref{comparison} compares the two variants of our method with the leading approaches in terms of AUC and processing frame rate. Our approach achieves competitive classification performance with the state-of-the-art at just a fraction of the computational cost. The cross-scene experiment, while inferior in terms of classification performance, is noteworthy in that each scene was classified using training data only from other surveillance scenarios.

\subsection{Violent-Flows Dataset}
The violent-flows dataset contains 246 clips containing violent (abnormal) and non-violent crowd behaviour. Classification is performed at the video level. A 5 fold cross validation evaluation approach is taken and results are compared in terms of mean accuracy. As both normal and abnormal training examples are available in this dataset our SVM-based classification approach is used. The majority classification found among the frames of a given clip is used as the overall result for that clip. An alternate approach is also taken where only the normal training examples are used and our GMM-based outlier detection approach is taken. Table \ref{GMM_VF} presents the BIC values calculated during the GMM selection stage, with z 4 component model ultimately used.  A full co-variance matrix GMM resulted in a lower BIC value in all cases and was therefore used. For this GMM-based approach the histogram of frame log probabilities for a given test clip is generated and the mode value is used to classify the overall clip by applying the Otsu threshold generated from the training data. Table \ref{bla} compares the two variations of our technique with the leading approaches in terms of mean accuracy and processing frame rate. Table \ref{contrib} highlights the contribution of each feature towards the achieved  anomaly detection accuracy on the violent-flows dataset using our SVM-based variant. Leaving out any individual feature results in a noticeable decrease in anomaly detection accuracy.

Our SVM-based variant achieves state-of-the-art performance on the violent-flows dataset with a mean accuracy of $85.53\pm0.17$\%. Our GMM-based variant achieves a very respectable $65.8\pm0.15$\% accuracy which is particularly impressive considering only half the training data, containing no violent behaviour, is used in this case. Our approach also achieves noticeably faster computational performance.

\begin{table}[tbp]
\centering
\begin{tabular}{|M{3cm}|M{2.5cm}|M{1.2cm}|}
\hline
Method & Accuracy & Speed (FPS) \\ \hline
SD \cite{Mohammadi2015}  & $85.43 \pm$0.21\% & N/A \\ 
HOT \cite{mousavi2015analyzing} & 82.3\% & N/A \\ 
ViF \cite{Hassneretal:SISM12} & $81.3\pm$0.18\% & 30 \\ 
CM \cite{mousavi2015crowd} & 81.5\% & 5 \\ \hline
Our Method (SVM) & \textbf{85.53$\pm$0.17\%} & \textbf{40} \\ 
Our Method (GMM) & $65.8\pm$0.15\% & \textbf{40} \\ \hline
\end{tabular}
\caption{Mean accuracy (with 95\% confidence interval where available) and processing speed on the violent-flows dataset.}
\label{bla}
\end{table}

\begin{table}[tbp]
\centering

\begin{tabular}{|M{3cm}|M{4cm}|}
\hline
Feature & Accuracy when excluded  \\ \hline
Crowd Collectiveness &$ 75.2\pm$0.1\%  \\ \hline
Crowd Conflict & $ 65.5\pm$0.16\%  \\ \hline
Crowd Density & $ 63.5\pm$0.13\% \\  \hline
Mean Motion Speed &$ 81.2\pm$0.12\%  \\  \hline
\end{tabular}
\caption{The contribution of each feature towards  mean detection accuracy on the violent-flows dataset using our SVM-based detection.}
\label{contrib}
\end{table}

\section{Conclusion}
This paper proposed a new set of features for crowd behaviour anomaly detection. These scene-level holistic features are easily interpretable, sensitive to abnormal crowd behaviour, and can be computed in better than real-time (40 frames per second) on commodity hardware. Our approach was demonstrated to improve upon the state-of-the-art classification performance on the violent-flows dataset.  Future work will attempt to improve upon certain limitations in our approach such as the scale issues present in our crowd density feature, possibly using an adaptive grid cell size. We also plan to  use this descriptor to label specific crowd behaviour concepts in larger and more challenging datasets.




\bibliographystyle{IEEEbib}
\bibliography{references}

\begin{thebibliography}{10}

\bibitem{zhou2014measuring}
Bolei Zhou, Xiaoou Tang, Hepeng Zhang, and Xiaogang Wang,
\newblock ``Measuring crowd collectiveness,''
\newblock {\em Pattern Analysis and Machine Intelligence, IEEE Transactions
  on}, vol. 36, no. 8, pp. 1586--1599, 2014.

\bibitem{Kang2015}
Jing Shao, Kai Kang, Chen~Change Loy, and Xiaogang Wang,
\newblock ``{Deeply Learned Attributes for Crowded Scene Understanding},''
\newblock in {\em Computer Vision and Pattern Recognition}, 2015.

\bibitem{Hassneretal:SISM12}
Y.~Itcher T.~Hassner and O.~Kliper-Gross,
\newblock ``Violent flows: Real-time detection of violent crowd behavior,''
\newblock in {\em 3rd IEEE International Workshop on Socially Intelligent
  Surveillance and Monitoring (SISM) at the IEEE Conf. on Computer Vision and
  Pattern Recognition (CVPR)}, 2012.

\bibitem{Li2014a}
Weixin Li, Vijay Mahadevan, and Nuno Vasconcelos,
\newblock ``{Anomaly detection and localization in crowded scenes},''
\newblock {\em IEEE Transactions on Pattern Analysis and Machine Intelligence},
  vol. 36, no. 1, pp. 18--32, 2014.

\bibitem{JavanRoshtkhari2013}
Mehrsan {Javan Roshtkhari} and Martin~D. Levine,
\newblock ``{An on-line, real-time learning method for detecting anomalies in
  videos using spatio-temporal compositions},''
\newblock {\em Computer Vision and Image Understanding}, vol. 117, no. 10, pp.
  1436--1452, 2013.

\bibitem{mehran2009abnormal}
Ramin Mehran, Akira Oyama, and Mubarak Shah,
\newblock ``Abnormal crowd behavior detection using social force model,''
\newblock in {\em Computer Vision and Pattern Recognition IEEE Conference on}.
  IEEE, 2009, pp. 935--942.

\bibitem{mousavi2015analyzing}
Hossein Mousavi, Sadegh Mohammadi, Alessandro Perina, Ryad Chellali, and
  Vittorio Murino,
\newblock ``Analyzing tracklets for the detection of abnormal crowd behavior,''
\newblock in {\em Applications of Computer Vision (WACV), 2015 IEEE Winter
  Conference on}. IEEE, 2015, pp. 148--155.

\bibitem{Mohammadi2015}
Sadegh Mohammadi, Alessandro Perina, and Istituto Italiano,
\newblock ``{Violence Detection in Crowded Scenes using Substantial
  Derivative},''
\newblock {\em Advanced Video and Signal Based Surveillance, IEEE Conference
  on}, 2015.

\bibitem{mousavi2015crowd}
Hossein Mousavi, M~Nabi, H~Kiani, A~Perina, and V~Murino,
\newblock ``Crowd motion monitoring using tracklet-based commotion measure,''
\newblock in {\em International Conference on Image Processing}. IEEE, 2015.

\bibitem{kaewtrakulpong2002improved}
Pakorn KaewTraKulPong and Richard Bowden,
\newblock ``An improved adaptive background mixture model for real-time
  tracking with shadow detection,''
\newblock in {\em Video-based surveillance systems}, pp. 135--144. Springer,
  2002.

\bibitem{shi1994good}
Jianbo Shi and Carlo Tomasi,
\newblock ``Good features to track,''
\newblock in {\em Computer Vision and Pattern Recognition, Proceedings, IEEE
  Computer Society Conference on}. IEEE, 1994, pp. 593--600.

\bibitem{tomasi1991detection}
Carlo Tomasi and Takeo Kanade,
\newblock {\em Detection and tracking of point features},
\newblock School of Computer Science, Carnegie Mellon Univ. Pittsburgh, 1991.

\bibitem{shao2014scene}
Jing Shao, Chen Loy, and Xiaogang Wang,
\newblock ``Scene-independent group profiling in crowd,''
\newblock pp. 2219--2226, 2014.

\bibitem{schwarz1978estimating}
Gideon Schwarz,
\newblock ``Estimating the dimension of a model,''
\newblock {\em The annals of statistics}, vol. 6, no. 2, pp. 461--464, 1978.

\bibitem{otsu1975threshold}
Nobuyuki Otsu,
\newblock ``A threshold selection method from gray-level histograms,''
\newblock {\em Automatica}, vol. 11, no. 285-296, pp. 23--27, 1975.

\end{thebibliography}

\end{document}